\documentclass{article}

\usepackage{times}
\usepackage{latexsym}

\usepackage{url}            
\usepackage{booktabs}       
\usepackage{amsfonts}       
\usepackage{nicefrac}       
\usepackage{microtype}      
\usepackage{natbib}
\usepackage{amsmath}
\usepackage{amsthm}
\usepackage{amssymb}
\usepackage{algorithm}
\usepackage{algorithmic}
\usepackage{enumitem}
\usepackage{defs}
\usepackage{graphicx}
\usepackage{caption}
\usepackage{color}
\usepackage{multirow}
\usepackage{sidecap}

\newcommand{\eda}{{\end{align}}}

\usepackage[accepted]{icml2017}

\title{A Neural Knowledge Language Model}

\icmltitlerunning{A Neural Knowledge Language Model}

\begin{document}

\twocolumn[
\icmltitle{A Neural Knowledge Language Model}



\icmlsetsymbol{equal}{*}

\begin{icmlauthorlist}
\icmlauthor{Sungjin Ahn}{UdeM}
\icmlauthor{Heeyoul Choi}{handong}
\icmlauthor{Tanel P\"arnamaa}{UdeM_intern}
\icmlauthor{Yoshua Bengio}{UdeM,cifar}
\end{icmlauthorlist}

\icmlaffiliation{UdeM}{Universit\'e de Montr\'eal, Canada}
\icmlaffiliation{UdeM_intern}{Work done during internship at the Universit\'e de Montr\'eal, Canada}
\icmlaffiliation{handong}{Handong Global University, South Korea}
\icmlaffiliation{cifar}{CIFAR Senior Fellow}

\icmlcorrespondingauthor{Sungjin Ahn}{sjn.ahn@gmail.com}


\icmlkeywords{boring formatting information, machine learning, ICML}

\vskip 0.3in
]

\printAffiliationsAndNotice{}  


\begin{abstract}
    Current language models have significant limitation in the ability to encode and decode factual knowledge. This is mainly because they acquire such knowledge from statistical co-occurrences although most of the knowledge words are rarely observed.~In this paper, we propose a Neural Knowledge Language Model (NKLM) which combines symbolic knowledge provided by the knowledge graph with the RNN language model.~By predicting whether the word to generate has an underlying fact or not, the model can generate such knowledge-related words by copying from the description of the predicted fact.~In experiments, we show that the NKLM significantly improves the performance while generating a much smaller number of unknown words.
\end{abstract}

\section{Introduction}

\begin{table}[h]
\centering
\scalebox{0.9}{
\textit{
\begin{tabular}{l}
Kanye West, a famous \textless unknown\textgreater { }and the husband of\\\textless unknown\textgreater, released his latest album \textless unknown\textgreater { } in\\\textless unknown\textgreater.
\end{tabular}}}
\label{my-label}
\end{table}
\vspace{-2mm}

A core purpose of language is to communicate knowledge.~For human-level language understanding, it is thus of primary importance for a language model to take advantage of knowledge.~Traditional language models are good at capturing statistical co-occurrences of entities as long as they are observed frequently in the corpus (e.g., words like verbs, pronouns, and prepositions).
However, they are in general limited in their ability in dealing with factual knowledge because these are usually represented by named entities such as person names, place names, years, etc. (as shown in the above example sentence of Kanye West.) 

Traditional language models have demonstrated to some extent the ability to encode and decode fatual knowledge~\citep{vinyals2015neural,serban2015hierarchical} when trained with a very large corpus.~However, we claim that simply feeding a larger corpus into a bigger model hardly results in a good knowledge language model.

The primary reason for this is the difficulty in learning good representations for rare and unknown words. This is a significant problem because these words are of our primary interest in knowledge-related applications such as question answering~\citep{iyyer2014neural,weston2015towards,bordes2015large} and dialogue modeling~\citep{vinyals2015neural,serban2015hierarchical}.~Specifically, in the recurrent neural network language model (RNNLM)~\citep{mikolov2010recurrent}, the computational complexity is linearly dependent on the number of vocabulary words.~Thus, including all words of a language is computationally prohibitive. Even if we can include a very large number of words in the vocabulary, according to Zipf's law, a large portion of the words will still be rarely observed in the corpus.


The fact that languages and knowledge can change over time also makes it difficult to simply rely on a large corpus.~Media produce an endless stream of new knowledge every day (e.g., the results of baseball games played yesterday) that is even changing over time.
Furthermore, a good language model should exercise some level of reasoning. For example, it may be possible to observe many occurrences of Barack Obama's year of birth and thus able to predict it in a correlated context. However, one would not expect current language models to predict, with a proper reasoning, the blank in ``\text{Barack Obama's age is \_\_\_\_}" even if it is only a simple reformulation of the knowledge on the year of birth\footnote{We do not investigate the reasoning ability in this paper but highlight this example because the explicit representation of facts would help to handle such examples.}. 

In this paper, we propose a Neural Knowledge Language Model (NKLM) as a step towards addressing the limitations of traditional language modeling when it comes to exploiting factual knowledge. In particular, we incorporate symbolic knowledge provided by the knowledge graph~\citep{nickel2015review} into the RNNLM. 
This connection makes sense particularly by observing that facts in knowledge graphs come along with textual representations which are mostly about rare words in text corpora.


In NKLM, we assume that each word generation is either based on a fact or not.~Thus, at each time step, before generating a word, we predict whether the word to generate has an underlying fact or not.~As a result, our model provides predictions over facts in addition to predictions over words.~Hence, the previous context information on both facts and words flow through an RNN and provide a richer context. The NKLM has two ways to generate a word.~One option is to generate a ``vocabulary word'' using the vocabulary softmax as is in the RNNLM. The other option is to generate a ``knowledge word'' by predicting the position of a word within the textual representation of the predicted fact. This makes it possible to generate words which are not in the predefined vocabulary and consequently resolves the rare and unknown word problem. The NKLM can also immediately adapt to adding or modifying knowledge because the model learns to predict facts, which can easily be modified without having to retrain the model.


The contributions of the paper are:
\bitem 
\item To propose the NKLM model to resolve limitations of traditional language models in dealing with factual knowledge by using the knowledge graph.
\item To develop a new dataset called \textit{WikiFact} which can be used in knowledge-related language models by providing text aligned with facts.
\item To show that the proposed model significantly improves the performance and can generate named entities which in traditional models were treated as unknown words.
\item To propose new evaluation metrics that resolve the problem of the traditional perplexity metric in dealing with unknown words.
\eitem 


\section{Related Work}

There have been remarkable recent advances in language modeling research based on neural networks~\citep{bengio2003neural,mikolov2010recurrent}.~In particular, the RNNLMs are interesting for their ability to take advantage of longer-term temporal dependencies without a strong conditional independence assumption.~It is especially noteworthy that the RNNLM using the Long Short-Term Memory (LSTM)~\citep{hochreiter1997long} has recently advanced to the level of outperforming carefully-tuned traditional n-gram based language models \citep{jozefowicz2016exploring}.

There have been many efforts to speed up the language models so that they can cover a larger vocabulary. These methods approximate the softmax output using hierarchical softmax~\citep{morin2005hierarchical,mnih2009scalable}, importance sampling~\citep{cho2015acl}, noise contrastive estimation~\citep{mnih2012fast}, etc. Although helpful to mitigate the computational problem, these approaches still suffer from the rare or unknown words problem. 

To help deal with the rare/unknown word problem, the pointer networks~\citep{Vinyals2015PointerN} have been adopted to implement the copy mechanism~\citep{gulcehre2016pointing,Gu2016IncorporatingCM} and applied to machine translation and text summarization. With this approach, the (unknown) word to copy from the context sentence is inferred from neighboring words. Similarly, \citet{merity2016pointer} proposed to copy from the context sentences and \citet{lebret2016neural} from Wikipedia infobox. However, because in our case the context can be very short and often contains no known relevant words (e.g., person names), we cannot use the existing approach directly.

Our knowledge memory is also related to the recent literature on neural networks with external memory~\citep{bahdanau2014neural,weston2014memory,graves2014neural}.~In \citet{weston2014memory}, given simple sentences as facts which are stored in the external memory, the question answering task is studied. In fact, the tasks that the knowledge-based language model aims to solve (i.e., predict the next word) can be considered as a fill-in-the-blank type of question answering.
The idea of jointly using Wikipedia and knowledge graphs has also been used in the context of enriching word embedding \citep{celikyilmaz2015enriching,long2016leveraging}. 

Context-dependent (or topic-based) language models have been studied to better capture long-term dependencies, by learning some context representation from the history. \citep{gildea1999topic} modeled the topic as a latent variable and proposed an EM-based approach. In \citep{mikolov2012context}, the topic features are learned by latent Dirichlet allocation (LDA)~\citep{blei2003latent}.

\section{Model}
\subsection{Preliminary}
A \textit{topic} is associated to \textit{topic knowledge} and \textit{topic description}.~Topic knowledge $\cF$ is a set of facts $\{a^{1},a^{2},\dots, a^{|\cF|}\}$ on the topic and topic description $W$ is a sequence of words $ (w_1, w_2, \dots, w_{|W|})$ describing the topic. We can obtain the topic knowledge from a knowledge graph such as Freebase and the topic description from Wikipedia.~In the corpus, we are given pairs of topic knowledge and topic description for $K$ topics, i.e.,  $\{(\cF_k, W_k)\}_{k=1}^K$. In the following, we omit index $k$ when we indicate an arbitrary topic.

A fact is represented as a triple of {\em subject}, {\em relationship}, and {\em object} which is associated with a textual representation, e.g., (\textit{Barack Obama, Married-To, Michelle Obama}). Note that all facts in a topic knowledge have the same subject entity which is the topic entity itself. 

We define \textit{knowledge words} $\cO_a$ of a fact $a$ as a sequence of words $(o_1^a, o_2^a, \dots, o_N^a)$ from which we can copy a word to generate output. We also maintain a global vocabulary $\cV$ containing frequent words. Because the words describing relationships (e.g., ``married to") are common and thus can be generated via the vocabulary $\cV$ not via copy, we limit the knowledge words of a fact to be the words for the object entity (e.g., $\cO_a$ = ($o_1^a$=``Michelle", $o_2^a$=``Obama"). In addition, to make it possible to access the subject words from the knowledge words, we add a special fact, (Topic, Topic\_Itself, Topic), to all topic knowledge. 


We train the model in a supervised way with labels on facts and words. This requires aligning words in the topic description with their corresponding facts in the topic knowledge. Specifically, given $\cF$ and $W$ for a topic, we perform simple string matching between the words in $W$ and all the knowledge words $\cO_\cF = \cup_{a\in\cF}\cO_{a}$ in such a way to associate fact $a$ to word $w$ if $w$ appears in knowledge words $\cO_a$. As a result, from $\cF$ and $W$, we construct a sequence of augmented observations $Y=\{y_t = (w_t,a_t,z_t)\}_{t=1:|W|}$. Here, $z_t$ is a binary variable indicating whether $w_t$ is observed in the knowledge words or not: 
\bea
z_t = 
\begin{cases}
    1,& \text{if } w_t \in \cO_{a_t},\\
    0,& \text{otherwise. } 
\end{cases}
\label{eq:z_t}
\eea

In addition, because not all words are associated to a fact (e.g., words like, {\em is, a, the, have}), we introduce a special fact type, called Not-a-Fact (NaF), and to which assign such words. The following is an example of an augmented observation induced from a topic description and knowledge.



\textbf{Example.}~Given a topic on Fred Rogers with topic description
\vspace{-2mm}
\begin{table}[h]
\centering
\scalebox{0.87}{
\textit{
\begin{tabular}{l}
$W$=``{\em Rogers was born in Latrobe, Pennsylvania in 1928}"
\end{tabular}}}
\label{my-label}
\end{table}
\vspace{-5mm}

and topic knowledge $\cF=\{a^{42},a^{83}, a^0\}$ where 
\vspace{-2mm}
\begin{table}[h]
\centering
\scalebox{0.87}{
\textit{
\begin{tabular}{l}
$a^{42}$ = (Fred\_Rogers, Place\_of\_Birth, Latrobe\_Pennsylvania)\\
$a^{83}$ = (Fred\_Rogers, Year\_of\_Birth, 1928)\\
$a^{0}$ \hspace{1.3mm}= (Fred\_Rogers, Topic\_Itself, Fred\_Rogers),
\end{tabular}}}
\label{my-label}
\end{table}

\vspace{-5mm}
the augmented observation $Y$ is
\vspace{-2mm}
\begin{table}[h]
\centering
\scalebox{0.87}{
\textit{
\begin{tabular}{l}
$Y$ = \{($w$=``Rogers", $a$=0, $z$=1), (``was", NaF, 0), \\
(``born", NaF, 0), (``in", NaF, 0), (``Latrobe", 42, 1), \\
(``Pennsylvania", 42, 1), (``in", NaF, 0), (``1928", 83, 1)\}.
\end{tabular}}}
\label{my-label}
\end{table}
\vspace{-2mm}



During inference and training of a topic, we assume that the topic knowledge $\cF$ is loaded in the \textit{knowledge memory} in a form of a matrix $\bF \in \eR^{D_a\times |\cF| }$ where the $i$-th column is a fact embedding $\baa^{i} \in \eR^{D_a}$. The fact embedding is the concatenation of subject, relationship, and object embeddings.~We obtain these entity embeddings from a preliminary run of a knowledge graph embedding method such as TransE~\citep{bordes2013translating}. Note that we fix the fact embedding during the training. Thus, there is no drift of fact embeddings after training and thus the model can deal with new facts at test time; we learn the embedding of the Topic\_Itself.

For notation, to denote the vector representation of an object of our interest, we use bold lowercase. For example, the embedding of a word $w$ is represented by $\bw = \bW[w]$ where $\bW^{D_w \times |\cV|}$ is the word embedding matrix, and $\bW[w]$ denotes the $w$-th column of $\bW$. 

\subsection{Inference}

\begin{figure}[t]
	\centering
	\vspace{-0.0cm}
		\includegraphics[width=0.35\textwidth]{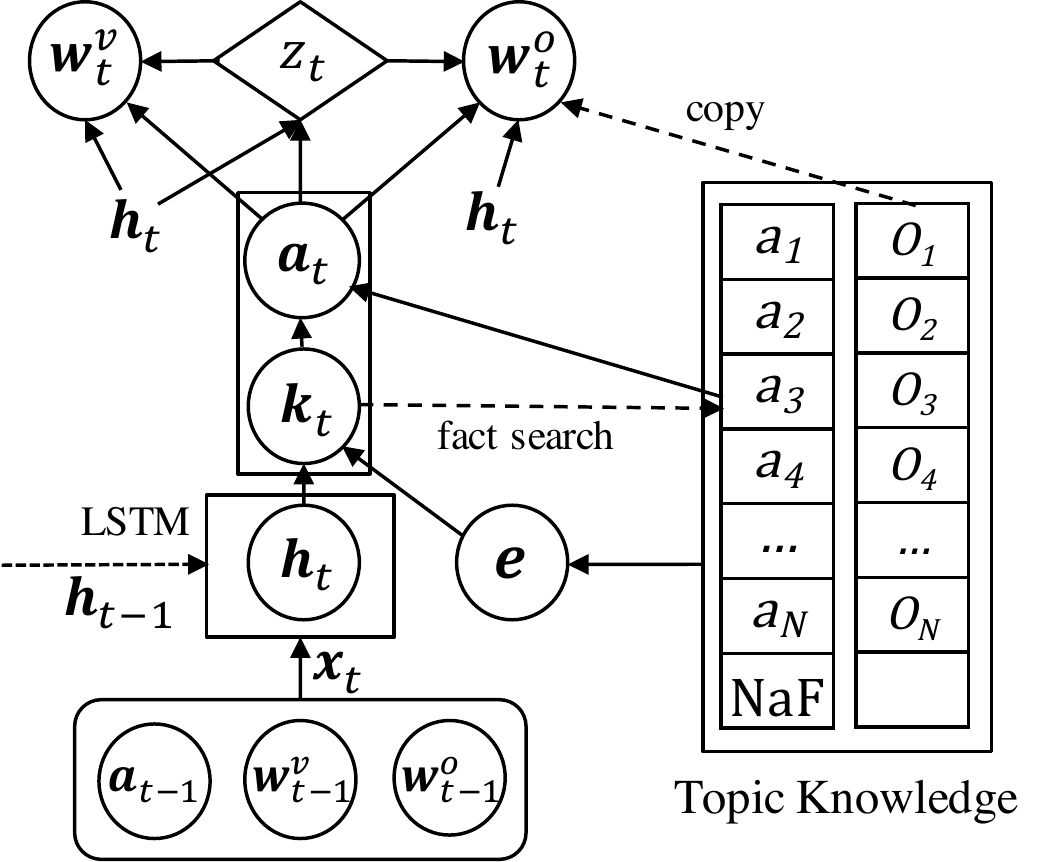}
\caption{The NKLM model. The input consisting of a word (either $\bw_{t-1}^o$ or $\bw_{t-1}^v$) and a fact ($\baa_{t-1}$) goes into LSTM. The LSTM's output $\bh_t$ together with the knowledge context $\bee$ generates the fact key $\bk_t$. Using the fact key, the fact embedding $\baa_t$ is retrieved from the topic knowledge memory.~Using $\baa_t$ and $\bh_t$, the word generation source $z_t$ is determined, which in turn determines the next word generation source $\bw_t^v$ or $\bw_t^o$. The copied word $\bw_t^o$ is a symbol taken from the fact description $\cO_{a_t}$.}
\label{fig:nklm_diag}
\end{figure}

At each time step, the NKLM performs the following four sub-steps: 
\vspace{-2mm}
\benum
\itemsep0em
\item Using both the word and fact predictions of the previous time step, make an input to the current time step and update the LSTM controller. 
\item Given the output of the LSTM controller, predict a fact and extract its corresponding embedding.
\item With the extracted fact embedding and the state of the LSTM controller, make a binary decision to determine the source of word generation. 
\item According to the chosen source, generate a word either from the global vocabulary or by copying a word from the knowledge words of the selected fact. 
\eenum
A model diagram is depicted in Fig. \ref{fig:nklm_diag}. In the following, we describe these steps in more detail. 
 

    \subsubsection{Input Representation and LSTM Controller} As shown in Fig.~\ref{fig:nklm_diag}, the input at time step $t$ is the concatenation of three embedding vectors corresponding to a fact $a_{t-1}$, a (global) vocabulary word $w_{t-1}^v \in \cV$, and a knowledge word $w_{t-1}^o \in \cO_{a_{t-1}}$, respectively. However, because the predicted word comes at a time step only either from the vocabulary or by copying from the knowledge words, i.e., $w_{t-1}\in\{w_{t-1}^v, w_{t-1}^o\}$ , we set either $\bw_{t-1}^v$ or $\bw_{t-1}^o$ to a zero vector when it is not the generation source at the previous step.
The resulting input representation $\bx_t = f_\text{concat}(\baa_{t-1}, \bw_{t-1}^v, \bw_{t-1}^o)$ is then fed into the LSTM controller, and obtain the output states $\bh_t= f_\text{LSTM}(\bx_{t}, \bh_{t-1})$. 
    \subsubsection{Fact Extraction} We then predict a relevant fact $a_t$ on which the word $w_t$ will be based. Predicting a fact is done in two steps.
    
First, a fact-key $\bk_\text{fact} \in \eR^{D_a}$ is generated by a function $f_\text{factkey}(\bh_t, \bee_k)$ which is in our experiments a multilayer perceptron (MLP) with one hidden layer of ReLU nonlinearity and linear outputs.~Here, $\bee_k \in \eR^{D_a}$ is the embedding of the topic knowledge which provides information about what facts are currently available in the topic knowledge. This would help the key generator adapt, without retraining, to changes in the topic knowledge such as removal or modification of some facts. 
Our experiments use mean-pooling to obtain $\bee_k$, but one can also consider using a more sophisticated method such as the soft-attention mechanism~\citep{bahdanau2014neural}. 

Then, using the generated fact-key $\bk_\text{fact}$, we select a fact by key-value lookup across the knowledge memory $\bF$ and then retrieve its embedding $\baa_t$ as follows:
\begin{align}
P(a|h_t) &= \f{\exp(\bk_\text{fact}\trns \bF[a])}{\sum_{a'\in \cF} \exp(\bk_\text{fact}\trns \bF[a'])},\\
 a_t &= \argmax_{a \in \cF} P(a|h_t),\\
\baa_t &= \bF[a_t].
\end{align}

    \subsubsection{Selecting Word Generation Source} Given the context $\bh_t$ and the extracted fact $\baa_t$, the model decides the source for the next word generation: either from the vocabulary $\cV$ or from the knowledge words $\cO_{a_t}$. We define the probability of selecting generation-by-copy as:
\begin{align}
\hat{z}_t = p(z_t|h_t,a_t) = \text{sigmoid}(f_\text{copy}(\bh_t, \baa_t)).
\end{align}
Here, $f_\text{copy}$ is an MLP with one ReLU hidden layer and a single linear output unit.

Word $w_t$ is generated from the source indicated by $\hat{z}_t$ as follows:
\bea
w_t = 
\begin{cases}
    w_t^v \in \cV,& \text{if } \hz_t < 0.5,\\
    w_t^o \in \cO_{a_t},& \text{otherwise. } 
\end{cases}\nonumber
\eea

\subsubsection{Word Generation} 

\textbf{Generation from Vocabulary Softmax:}
For vocabulary word $w_t^v \in \cV$, we follow the usual way of selecting a word using the softmax function:
\begin{align}
P(w_t^v = w |h_t,a_t) &= \f{\exp(\bk_\text{voca}\trns \bW[w])}{\sum_{w' \in \cV}\exp(\bk_\text{voca}\trns \bW[w'])},
\end{align}

where $\bk_\text{voca} \in \eR^{D_w}$ is obtained by $f_\text{voca}(\bh_t , \baa_t)$ which is an MLP with a ReLU hidden layer and $D_w$ linear output units.

\textbf{Generation by Copy from Knowledge Words:}
To copy a knowledge word $w_t^o \in \cO_{a_t}$, we first predict the \emph{position} of the word within the knowledge words and then copy the word on the predicted position. This copy-by-position allows us not to rely on the word embeddings by instead learning position embeddings.

\begin{algorithm}[t]
\caption{NKLM inference at time step $t$}\label{alg:nklm}
\begin{algorithmic}[1]
\STATE \#\# \textit{Make input} 
\STATE $\bx_t = f_\text{concat}(\baa_{t-1}, \bw_{t-1}^v, \bw_{t-1}^o)$

\STATE \#\# \textit{Update LSTM controller}
\STATE $\bh_t = \text{LSTM}(\bh_{t-1}, \bx_t)$\\
\STATE \#\# \textit{Fact prediction and extract embedding}
\STATE $a_t = \argmax_{a\in \cF} P(a|\bh_t, \bee_k)$ 
\STATE $\baa_t = \bF[a_t]$
\STATE \#\# \textit{Decide word generation source}
\STATE $z_t = \eI[p(z_t|\bh_t,\baa_t) > 0.5]$
\IF {$z_t == 0$ } 
\STATE \#\# \textit{Word generation from vocabulary}
\STATE $w_t = w_t^v = \argmax_{w \in \cV} P(w|\bh_t,\baa_t)$
\STATE $\bw_t^o = \bzero$
\ELSE
\STATE \#\# \textit{Word generation by copy}
\STATE $n_t = \argmax_{n=0:|\cO_{a_t}|-1} P(n|\bh_t,\baa_t) $
\STATE $w_t = w_t^o = \cO_{a_t}[n_t]$
\STATE $\bw_t^v = \bzero$
\ENDIF
\end{algorithmic}
\end{algorithm}

\begin{table*}[t]
	\centering
	\scalebox{0.97}{
    \begin{tabular}{c | c | c | c |c | c | c | c | c | c }
\# topics  &  \# toks  & \#  uniq toks & \# facts & \# entities & \# relations & $\max_k |\cF_k|$ & $\text{avg}_k |\cF_k|$ & $\max_a|O_{a}|$ & $\text{avg}_a |\cO_a|$\\ 
     \hline\hline 
10K & 1.5M &  78k &  813k & 560K & 1.5K & 1K & 79 & 19 & 2.15
    	\end{tabular}}
	\caption{Statistics of the WikiFacts-FilmActor-v0.1 dataset.}	\label{tb:stat}	
\end{table*}

One reason to use position prediction is that the traditional copy mechanism~\citep{gulcehre2016pointing,Gu2016IncorporatingCM} is difficult to apply to our context because the knowledge words usually consist of only unknown words and/or are short in length. Furthermore, it makes sense when considering the fact that we mostly need to copy the knowledge words in increasing order from the first word. For example, given that the first symbol $o_{1} = \textit{``Michelle"}$ was used in the previous time step and prior to that other words such as \textit{``President''} and {\em ``US''} were also observed, the model can easily predict that it is time to select the second symbol, i.e., $o_2=\textit{``Obama"}$. 

More specifically, we first generate the position key $\bk_\text{pos} \in \eR^{D_o}$ by a function $f_\text{poskey}(\bh_t, \baa_t)$ which is again an MLP with one hidden layer and linear outputs whose dimension is the maximum number of positions, e.g., the maximum length of the knowledge words (e.g., $N^o_\text{max} = \max_{a \in \bar{\cF}} |O_{a}|$ where $\bar{\cF} = \cup_k \cF_k$). Then, the word to copy is chosen by
\begin{align}
    P(n |h_t, a_t) &= \f{\exp(\bk_\text{pos}\trns \bP[n])}{\sum_{n'}\exp(\bk_\text{pos}\trns \bP[n'])},\\
    n_t &= \argmax_{n=0:|\cO_{a_t}|-1} P(n |h_t, a_t),\\
    w_t^o &= \cO_{a_t}[n_t],
\end{align}
with position $n'$ running from 0 to $|\cO_{a_t}|-1$.
Here, $\bP^{D_o \times N^o_\text{max}}$ is the matrix of position embeddings of dimension $D_o$. Note that $N^o_\text{max}$ is typically a much smaller number (e.g., 20 in our experiments) than the size of vocabulary, and thus the computation for copy is efficient. The position embedding matrix $\bP$ is learned during training.

Although in our experiments  we find that the simple position prediction performs well, we note that one could also consider a more advanced encoding such as one based on a convolutional network~\citep{kim2014convolutional} to model the knowledge words. 

To compute $p(w_t | w_{<t}, \cF)$, we first obtain $\{z_{<t}, a_{<t}\}$ from $\{w_{<t}\}$ and $\cF$ using the augmentation procedure, and perform the above inference process with hard decisions taken about $z_t$ and $a_t$ based on the model's predictions. The inference procedure is summarized in Algorithm \ref{alg:nklm}.

\subsection{Learning}
We perform supervised learning on the augmented observation $Y$, similarly to \citet{reed2015neural}. That is, given word observations $\{Y_k\}_{k=1}^K$ and knowledge $\{\cF_k\}_{k=1}^K$, our objective is to maximize the log-likelihood of the augmented observation w.r.t the model parameter $\ta$,
\begin{align}
\ta^* = \argmax_\ta \sum_{k} \log P_\ta(Y_k|\cF_k).
\end{align}
By the chain rule, we can decompose the probability of the observation $Y_k$ as
\begin{align}
\log P_\ta(Y_k |\cF_k) = \sum_{t=1}^{|Y_k|} \log P_\ta(y_t^k | y_{1:t-1}^k, \cF_k).
\end{align}
Then, after omitting $\cF_k$ and $k$ for simplicity, we can rewrite the single step conditional probability as
\begin{align}
    P_\ta(y_t|y_{1:t-1}) &= P_\ta(w_t, a_t, z_t|h_t)\\
&= P_\ta(w_t | a_t, z_t, h_t) P_\ta(a_t|h_t) P_\ta(z_t|h_t).\nn\label{eqn:wt}
\end{align}
We maximize the above objective using stochastic gradient optimization.

\begin{table*}[t]
\centering
\scalebox{1.0}{
\begin{tabular}{l||rrr|rrr|r}
   \multirow{1}{*}{} &
      \multicolumn{3}{c|}{Validation} &
      \multicolumn{3}{c|}{Test} \\
  \hline
Model &  PPL & UPP & UPP-f & PPL & UPP & UPP-f & \# UNK \\ 
  \hline\hline
  RNNLM &  39.4 & 97.9 & 56.8 & 39.4 & 107.0 & 58.4 & 23247 \\ 
  \hline
  NKLM & \bf{27.5} & \bf{45.4} & \bf{33.5} & \bf{28.0} & \bf{48.7} & \bf{34.6} & \bf{12523} \\
  no-copy &  38.4	& 93.5	& 54.9	& 38.3	& 102.1 &	56.4 & 29756\\ 
  no-fact-no-copy & 40.5	& 98.8	& 58.0	& 40.3	& 107.4	& 59.3  & 32671\\
  no-TransE &  48.9	& 80.7	& 59.6	& 49.3	& 85.8	& 61.0 & {13903} \\ 
   \hline
\end{tabular}}
\caption{\textbf{We compare four different versions of the NKLM to the RNNLM on three different perplexity metrics.} We used 10K vocabulary. In \textbf{no-copy}, we disabled the generation-by-copy functionality, and in \textbf{no-fact-no-copy}, using topic knowledge is also additionally disabled by setting all facts as NaF. Thus, \textbf{no-fact-no-copy} is very similar to RNNLM. In \textbf{no-TransE}, we used random vectors instead of the TransE embeddings to initialize the knowledge graph entities. As shown, the NKLM shows best performance in all cases. The \textbf{no-fact-no-copy} performs similar to the RNNLM as expected (slightly worse partly because it has a smaller number of model parameters than that of the RNNLM). As expected, \textbf{no-copy} performs better than \textbf{no-fact-no-copy} by using additional information from the fact embedding, but without the copy mechanism. In the comparison of the NKLM and \textbf{no-copy}, we can see the significant gain of using the copy mechanism to predict named entities. In the last column, we can also see that, with the copy mechanism, the number of predicting unknown decreases significantly. Lastly, we can see that the TransE embedding is important.}
\label{tbl:perp_main}
\end{table*}

\begin{table*}[t]
\centering
\scalebox{1.0}{
\begin{tabular}{l||rrr|rrr|r}
   \multirow{1}{*}{} &
      \multicolumn{3}{c|}{Validation} &
      \multicolumn{3}{c|}{Test} \\
  \hline
Model & PPL & UPP & UPP-f & PPL & UPP & UPP-f & \# UNK \\ 
  \hline\hline
NKLM\_5k & \textbf{22.8} & \textbf{48.5} & \textbf{30.7} & \textbf{23.2} & \textbf{52.0} & \textbf{31.7} & \textbf{19557} \\
RNNLM\_5k &  27.4 & 108.5 & 47.6 & 27.5 & 118.3 & 48.9 & 34994 \\ \hline
  NKLM\_10k & \bf{27.5} & \bf{45.4} & \bf{33.5} & \bf{28.0} & \bf{48.7} & \bf{34.6} & \bf{12523} \\ 
  RNNLM\_10k &  39.4 & 97.9 & 56.8 & 39.4 & 107.0 & 58.4 & 23247 \\ \hline
  NKLM\_20k & \textbf{33.4} & \textbf{45.9} & \textbf{37.9} & \textbf{34.7} & \textbf{49.2} & \textbf{39.7} & \textbf{9677} \\ 
  RNNLM\_20k &  57.9 & 99.5 & 72.1 & 59.3 & 108.3 & 75.5 & 13773 \\ \hline
  NKLM\_40k & \textbf{41.4} & \textbf{49.0} & \textbf{44.4} & \textbf{43.6} & \textbf{52.7} & \textbf{47.1} & \textbf{5809} \\ 
  RNNLM\_40k & 82.4 & 107.9 & 92.3 & 86.4 & 116.9 & 97.9 & 9009 \\
   \hline
\end{tabular}}
\caption{\textbf{The NKLM and the RNNLM are compared for vocabularies of four different sizes [5K, 10K, 20K, 40K]}. As shown, in all cases the NKLM significantly outperforms the RNNLM. Interestingly, for the standard perplexity (PPL), the gap between the two models increases as the vocabulary size increases while for UPP the gap stays at a similar level regardless of the vocabulary size. This tells us that the standard perplexity is significantly affected by the UNK predictions, because with UPP the contribution of UNK predictions to the total perplexity is very small. Also, from the UPP value for the RNNLM, we can see that it initially improves when vocabulary size is increased as it can cover more words, but decreases back when the vocabulary size is largest (40K) because the rare words are added last to the vocabulary.}
\label{tbl:perp_voca_sz}
\end{table*}

\begin{table*}[t]
\centering
\resizebox{\textwidth}{!}{%
\begin{tabular}{l|l}
\hline\hline
Warm-up &  Louise Allbritton ( 3 july \textless unk\textgreater february 1979 ) was\\ 
\hline
RNNLM &  a \textless unk\textgreater \textless unk\textgreater who was born in \textless unk\textgreater , \textless unk\textgreater , \textless unk\textgreater , \textless unk\textgreater , \textless unk\textgreater , \textless unk\textgreater , \textless unk\textgreater\\ 
NKLM & an english [Actor]. he was born in [Oklahoma] , and died in [Oklahoma]. he was married to [Charles] [Collingwood]\\ 
   \hline\hline
Warm-up &  Issa Serge Coelo ( born 1967 ) is a \textless unk\textgreater\\ 
\hline
RNNLM &  actor . he is best known for his role as \textless unk\textgreater \textless unk\textgreater in the television series \textless unk\textgreater . he also\\ 
NKLM & [Film] director . he is best known for his role as the \textless unk\textgreater \textless unk\textgreater in the film [Un] [taxi] [pour] [Aouzou]\\ 
   \hline\hline
Warm-up &  Adam wade Gontier is a canadian Musician and Songwriter .\\ 
\hline
RNNLM &  she is best known for her role as \textless unk\textgreater \textless unk\textgreater on the television series \textless unk\textgreater . she has also appeared\\ 
NKLM &  he is best known for his work with the band [Three] [Days] [Grace] . he is the founder of the
\\
\hline\hline
Warm-up &  Rory Calhoun ( august 8 , 1922 – april 28 \\ 
\hline
RNNLM &  , 2010 ) was a \textless unk\textgreater actress . she was born in \textless unk\textgreater , \textless unk\textgreater , \textless unk\textgreater . she was\\ 
NKLM &  , 2008 ) was an american [Actor] . he was born in [Los] [Angeles] california . he was born in
\\
\hline\hline 
\end{tabular}}\caption{\textbf{Sampled Descriptions}. Given the warm-up phrases, we generate samples from the NKLM and the RNNLM. We denote the copied knowledge words by [word] and the UNK words by \textless unk\textgreater. Overall, the RNNLM generates many UNKs (we used 10K vocabulary) while the NKLM is capable to generate named entities even if the model has not seen some of the words at all during training. In the first case, we found that the generated symbols (words in []) conform to the facts of the topic (Louise Allbritton) except that she actually died in Mexico, not in Oklahoma. (We found that the place\_of\_death fact was missing.) While she is an actress, the model generated a word [Actor]. This is because in Freebase, there exists only /profession/actor but no /profession/actress. It is also noteworthy that the NKLM fails to use the gender information provided by facts; the NKLM uses ``he'' instead of ``she'' although the fact /gender/female is available. From this, we see that if a fact is not detected (i.e., NaF), the statistical co-occurrence governs the information flow. Similarly, in other samples, the NKLM generates movie titles (Un Taxi Pour Aouzou), band name (Three Days Grace), and place of birth (Los Angeles). In addition, to see the NKLM's ability to adapt to knowledge updates without retraining, we changed the fact /place\_of\_birth/Oklahoma to /place\_of\_birth/Chicago and found that the NKLM replaces ``Oklahoma" by ``Chicago" while keeping other words the same.}
\label{tbl:samples}
\end{table*}

\section{Evaluation}

An obstacle in developing the proposed model is the lack of datasets where the text is aligned with facts at the word level. While the Penn Treebank (PTB) dataset \citep{ptb_dataset} has been frequently used in language modeling, as pointed by \citet{merity2016pointer}, its limited vocabulary containing a relatively small amount of named entities makes it difficult to use them for knowledge-related tasks where rare words are of primary interest; we would have only a very small amount of words to be associated with facts. As other larger datasets such as in \citet{chelba2013one} also have problems in licensing or in the format of the dataset, we produce the \textit{WikiFacts} dataset for evaluation of the proposed model and the baseline model. The dataset is freely available in {\url{https://bitbucket.org/skaasj/wikifact_filmactor}}.

\subsection{The WikiFacts Dataset}
In \textit{WikiFacts}, we align Wikipedia descriptions with corresponding Freebase\footnote{Freebase has migrated to Wikidata. \url{www.wikidata.org}} facts.~Because many Freebase topics provide a link to its corresponding topic in Wikipedia, we choose a set of topics for which both a Freebase entity and a Wikipedia description exist. In the experiments, we used a version called \texttt{WikiFacts-FilmActor-v0.1} where the domain is restricted to the \textit{/Film/Actor} in Freebase. 


We used the summary part (first few paragraphs) of the Wikipedia page as the text to be modeled, but discarded topics for which the number of facts is too large ($>$ 1000) or the Wikipedia description is too short ($<$ 3 sentences). For the string matching, we also used synonyms and alias information provided by WordNet~\citep{miller1995wordnet} and Freebase. 

We augmented the fact set $\cF$ with the \textit{anchor} facts $\cA$ whose relationship is all set to {\tt UnknownRelation}.~That is, observing that an anchor (a word under a hyperlink) in a Wikipedia description has a corresponding Freebase entity as well as being semantically closely related to the topic in which the anchor is found, we make a synthetic fact of the form (\textit{Topic}, {\tt UnknownRelation}, \textit{Anchor}). This potentially compensates for some missing facts in Freebase. Because we extract the anchor facts from the full Wikipedia page and they all share the same relation, it is more challenging for the model to use these anchor facts than using the Freebase facts.

As a result, for each word $w$ in the description, we obtain a tuple ($w, z, a, n, k$). Here, $w$ is word id, $z$ the copy indicator, $a$ fact id, $n$ the position to copy from $\cO_a$ if $z=1$, and $k$ topic id. We provide a summary of the dataset statistics in Table~\ref{tb:stat}. 



\subsection{Experiments}

\subsubsection{Setup} 
We split the dataset into 80/10/10 for train, validation, and test. As a baseline model, we use the RNNLM. For both the NKLM and the RNNLM, two-layer LSTMs with dropout regularization~\citep{zaremba2014recurrent} are used. We tested models with different numbers of LSTM hidden units [200, 500, 1000], and report results from the 1000 hidden-unit model.
For the NKLM, we set the symbol embedding dimension to 40 and word embedding dimension to 400. Under this setting, the number of parameters in the NKLM is  slightly smaller than that of the RNNLM. 

We used 100-dimension TransE embeddings for Freebase entities and relations, and concatenate the relation and object embeddings to obtain fact embeddings. We averaged all fact embeddings in $\cF_k$ to obtain the topic knowledge embedding $\bee_k$.
We unrolled the LSTMs for 30 steps and used minibatch size 20. We trained the models using stochastic gradient ascent with gradient clipping range [-5,5]. The initial learning rate was set to 0.5 for the NKLM and 1.5 for the RNNLM, and decayed after every epoch by a factor of 0.98. We trained for 50 epochs and report the results chosen by the best validation set results. 

\subsubsection{The Unknown Penalized Perplexity }
The perplexity $\exp(-\f{1}{N}\sm{i}{N}\log p(w_i))$ is the standard performance metric for language modeling.~This, however, has a problem in evaluating language models for a corpus containing many named entities: \emph{a model can get good perplexity by accurately predicting UNK words as the UNK class}. As an extreme example, when all words in a sentence are unknown words, a model predicting everything as UNK will get a good perplexity.~Considering that unknown words provide virtually no useful information, this is clearly a problem in tasks where named entities are important such as question answering, dialogue modeling, and knowledge language modeling. 

To this end, we propose a new evaluation metric, called the Unknown-Penalized Perplexity (UPP), and evaluate the models on this metric as well as the standard perplexity (PPL). Because the actual word underlying the UNK should be one of the out-of-vocabulary (OOV) words, in UPP we penalize the likelihood of unknown words as follows:
$$P_{\text{UPP}}(w_\text{unk}) = \frac{P(w_\text{unk})}{ |\cV_\text{total} \setminus \cV_\text{voca}|}.$$
Here, $\cV_\text{total}$ is a set of all unique words in the corpus, and $\cV_\text{voca} \subset \cV_\text{total}$ is the global vocabulary used for word generation. In other words, in UPP we assume that the OOV set is equal to $\cV_\text{total} \setminus \cV_\text{voca}$ and thus assign a uniform probability to OOV words. 
In another version, UPP-fact, we consider the fact that the RNNLM can also use the knowledge given to the NKLM to some extent, but with limited capability (because the model is not designed for it). For this, we assume that the OOV set is equal to the total knowledge words of a topic $k$, i.e., 
$$P_{\text{UPP-fact}}(w_\text{unk}) = \frac{P(w_\text{unk})}{|\cO_{\cF_k}|},$$
where $\cO_{\cF_k} = \cup_{a\in \cF_k} \cO_{a}$. In other words, by using UPP-fact, we assume that, for an unknown word, the RNNLM can pick one of the knowledge words with uniform probability. 

\subsubsection{Observations from experiment results} 
We describe the detail results and analysis on the experiments in detail in the captions of Table~\ref{tbl:perp_main}, ~\ref{tbl:perp_voca_sz}, and ~\ref{tbl:samples}. Our observations from the experiment results are as follows.
\bitem 
\itemsep0em
    \item The NKLM outperforms the RNNLM in all three perplexity measures. 
    \item The copy mechanism is the key of the significant performance improvement. Without the copy mechanism, the NKLM still performs better than the RNNLM due to its usage of the fact information, but the improvement is not so significant.~
    \item The NKLM results in a much smaller number of UNKs (roughly, a half of the RNNLM). 
    \item When no knowledge is available, the NKLM performs as well as the RNNLM. 
    \item Knowledge graph embedding using TransE is an efficient way of representing facts in our model.
    \item The NKLM generates named entities in the provided facts whereas the RNNLM generates many more UNKs. 
    \item The NKLM shows its ability to adapt immediately to the change of the knowledge. 
    \item The standard perplexity is significantly affected by the prediction accuracy on the unknown words. Thus, one need carefully consider when using it as a metric for knowledge-related language models. 
\eitem 



\section{Conclusion}
In this paper, we presented a novel Neural Knowledge Language Model (NKLM) that brings the symbolic knowledge from a knowledge graph into the expressive power of RNN language models.~The NKLM significantly outperforms the RNNLM in terms of perplexity and generates named entities which are not observed during training, as well as immediately adapting to changes in knowledge. We believe that the WikiFact dataset introduced in this paper, can be useful in other knowledge-related language tasks as well. In addition, the Unknown-Penalized Perplexity introduced in order to resolve the limitation of the standard perplexity, can  also be useful in evaluating other language tasks. 

The task that we investigated in this paper is limited in the sense that we assume that the true topic of a given description is known.~Relaxing this assumption by making the model search for a proper topic on-the-fly will make the model more practical and scalable. We believe that there are many more open research challenges related to the knowledge language models.

\section*{Acknowledgments}
The authors would like to thank Alberto Garc\'ia-Dur\'an, Caglar Gulcehre, Chinnadhurai Sankar, Iulian Serban, Sarath Chandar, and Peter Clark for helpful feedbacks and discussions as well as the developers
of Theano~\citep{bastien2012theano}, NSERC, CIFAR, Facebook, Google, IBM, Microsoft, Samsung, and Canada Research Chairs for funding, and Compute Canada for computing resources.

\bibliography{./nklm}
\bibliographystyle{icml2017}

\newpage
\onecolumn
\subsubsection*{Appendix: Heatmaps}
\begin{figure*}[ht]
	\centering
	\vspace{-0.5cm}
		\includegraphics[width=0.9\textwidth]{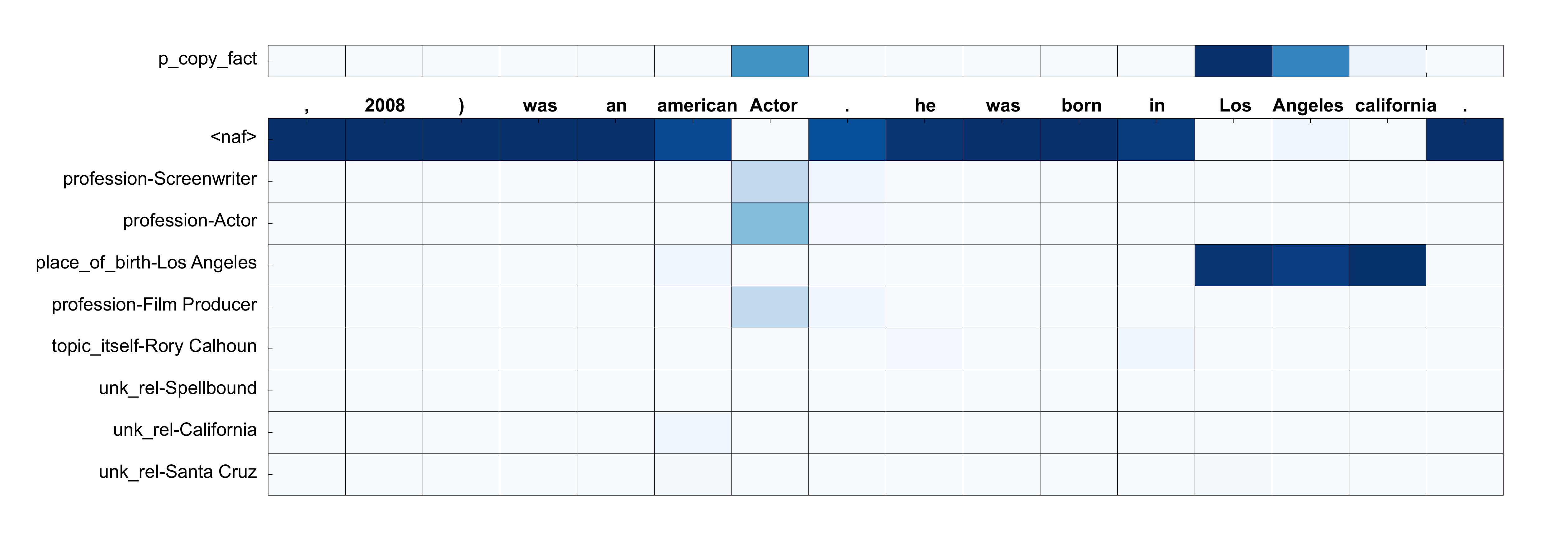}
    \vspace{-0.5cm}
\caption{This is a heatmap of an example sentence generated by the NKLM having a warmup {\em``Rory Calhoun ( august 8 , 1922 april 28''}. The first row shows the probability of selecting copy (Equation 5 in Section 3.1). The bottom heat map shows the state of the topic-memory at each time step (Equation 2 in Section 3.1). In particular, this topic has 8 facts and an additional \textless NaF\textgreater{ }fact. For the first six time steps, the model retrieves \textless NaF\textgreater from the knowledge memory, copy-switch is off and the words are generated from the general vocabulary. For the next time step, the model gives higher probability to three different profession facts: ``Screenwriter'', ``Actor'' and ``Film Producer.'' The fact ``Actor'' has the highest probability, copy-switch is higher than 0.5, and therefore ``Actor'' is copied as the next word. Moreover, we see that the model correctly retrieves the place of birth fact and outputs ``Los Angeles.'' After that, the model still predicts the place of birth fact, but copy-switch decides that the next word should come from the general vocabulary, and outputs ``California.''}
\label{fig:nklm}
\end{figure*}

\begin{figure*}[ht]
	\centering
	\vspace{-0.5cm}
		\includegraphics[width=0.9\textwidth]{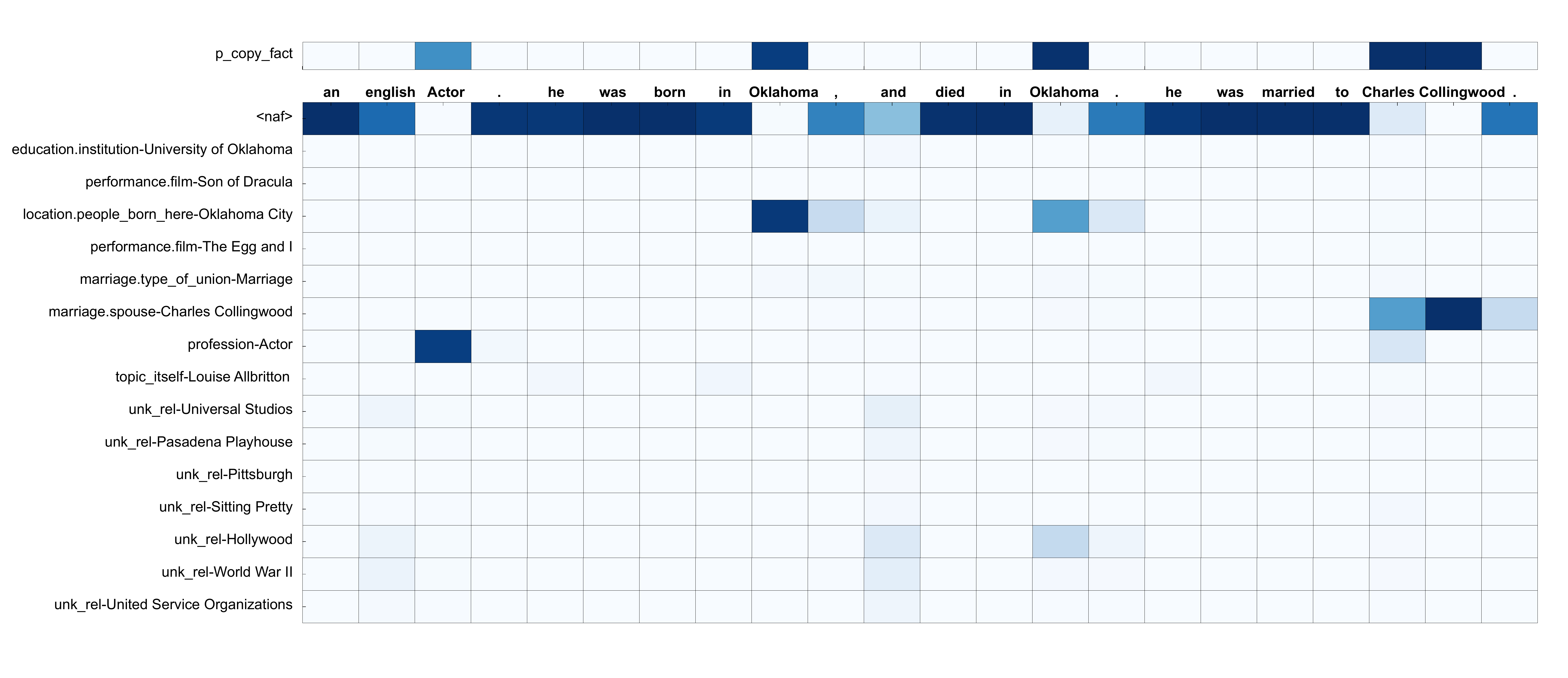}
	\vspace{-0.5cm}
\caption{This is an example sentence generated by the NKLM having a warmup {\em``Louise Allbritton ( 3 july \textless unk\textgreater  february 1979 ) was''}. We see that the model correctly retrieves and outputs the profession (``Actor''), place of birth (``Oklahoma''), and spouse (``Charles Collingwood'') facts. However, the model makes a mistake by retrieving the place of birth fact in a place where the place of death fact is supposed to be used. This is probably because the place of death fact is missing in this topic memory and then the model searches for a fact about location, which is somewhat encoded in the place of birth fact. In addition, {\em Louise Allbritton} was a woman, but the model generates a male profession ``Actor'' and male pronoun ``he''. The ``Actor'' is generated because there is no ``Actress'' representation in Freebase.}
\label{fig:nklm}
\end{figure*}

\end{document}